\documentclass[runningheads]{llncs}
\usepackage{hyperref}
\usepackage{amsmath}
\usepackage{graphicx}
\usepackage{enumitem}
\usepackage{amsmath,amssymb} 
\usepackage{color}
\usepackage[width=122mm,left=12mm,paperwidth=146mm,height=193mm,top=12mm,paperheight=217mm]{geometry}

\usepackage{algpseudocode} 
\usepackage{amsmath}
\usepackage{amsfonts}
\usepackage{bm}
\usepackage{enumerate}
\usepackage{verbatim}
\usepackage{array} 
\usepackage{booktabs} 
\usepackage{xspace}
\usepackage{graphicx}

\newcommand{\paracompact}{\noindent\textbf}

\newcommand{\beginsupplement}{%
\setcounter{section}{0}
\renewcommand{\thesection}{\Alph{section}}
\renewcommand{\theHsection}{appendixsection.\Alph{section}}
\setcounter{table}{0}
\renewcommand{\thetable}{S\arabic{table}}%
\setcounter{figure}{0}
\renewcommand{\thefigure}{S\arabic{figure}}%
}

\begin{document}
\pagestyle{headings}
\mainmatter

\title{Human Motion Modeling using DVGANs} 

\titlerunning{Human Motion Modeling using DVGANs}

\authorrunning{Xiao Lin and Mohamed R. Amer}

\author{Xiao Lin and Mohamed R. Amer}
\institute{SRI International\\Princeton, NJ08540, USA\\\email{firstname.lastname@sri.com}}
\maketitle
\begin{abstract}
We present a novel generative model for human motion modeling using Generative Adversarial Networks (GANs). We formulate the GAN discriminator using dense validation at each time-scale and perturb the discriminator input to make it translation invariant. Our model is capable of motion generation and completion. We show through our evaluations the resiliency to noise, generalization over actions, and generation of long diverse sequences. We evaluate our approach on Human 3.6M and CMU motion capture datasets using inception scores.

\end{abstract}
\section{Introduction}
Modeling human motion is an important problem in vision and graphics. The applications cover interactions in virtual and physical worlds, robotics, animation, and surveillance. The tasks vary between classification, retrieval, synthesis, and completion. Human motion is a rich spatio-temporal representation, with many variations, which requires efficient modeling. It is highly structured since the body joints depend on each other per frame, and their motion across frames.

Learning a rich representation would enable all of the tasks above. With recent advances in deep learning with applications in vision, language, and machine translation fields, learning rich representations became possible. In combination with large scale human motion datasets such as Human 3.6M \cite{h36m_pami}, human motion modeling witnessed a large boost in performance. Due to the focus on motion completion, generative, probabilistic, and time-series models are necessary to model human motion. Early work considered Restricted Boltzmann Machines (RBMs) for motion generation \cite{Taylor_JMLR2011}. Recent approaches successfully used variations of Recurrent Neural Networks (RNNs) \cite{FragkiadakiLM15,JainZSS16,GhoshSAH17} or sequence-to-sequence models \cite{MartinezBR17,LiZXHL18} addressing the motion completion task on Human 3.6M dataset. While the focus is always on motion completion, not many approaches focused on motion generation from scratch, there was always required a seed motion to initialize the model. In departure from recent work, our model's main focus is primarily on motion generation and we also are able to do motion completion. 

With the recent successes of GANs \cite{Goodfellow14}, a generative framework that model implicit densities, in realistic image generation \cite{RadfordMC15}, completion \cite{IizukaSSI17}, translation \cite{ZhuPIE17,LiuBK17}, manipulation \cite{WangLZTKC17}, and super resolution \cite{LedigTHCAATWS17}. These successes makes GANs a great candidate for human motion modeling for generation and completion. Even with these successes in the image domain, GANs has not been applied in human motion modeling as much \cite{AhnHCYO17,BarsoumBKL17}. This could be due to its complexity in training its RNN generator and discriminator. We propose a novel model, Dense Validation GANs (DVGANs) model for motion generation and completion. DVGANs uses combination of convolutional and recurrent discriminators and generators. It treats class labels as text which turns out to be a good regularizer for our model. This would enable us to extend our model in the future to accept free-form text and generate smooth compositional action sequences.

Evaluating generative models is an open research problem, the generative modeling community has developed various ad-hoc evaluative criteria. For human motion completion, \cite{FragkiadakiLM15} proposed Euclidean norm between the completed motion sequence and ground-truth motion. While this metric is successful at measuring the model's ability to generate data as close as possible to the ground truth data, it fails with longer sequences (larger than 560mseconds), thus most work resort to qualitative comparisons. Another metric is inception score, which has been a popular method to evaluate image generation quality and diversity \cite{RadfordMC15}. Since our interest is in motion generation from scratch, focusing on how diverse and realistic the samples are is more important to us than measuring how close the generated samples to the groundtruth. Meaning that if our model is able to generalize and learn how to combine motion successfully from the different actions, yet preserve the overall motion of the specific action it should not be penalized. We rigorously investigate the different metrics and benchmark our results using the Inception Score. We are also the first to benchmark the whole CMU Mocap Database \cite{CMUMocap} for motion generation. \noindent{\it Our Contributions Summary:}
\begin{itemize}[noitemsep,topsep=-1pt,leftmargin=15pt]
	\item A novel DVGANs model for human motion generation.
	\item Conditioning on text rather than class labels.
	\item Generating long, high quality animations.
	\item Benchmarking on the full CMU Motion Capture database.
\end{itemize}
\noindent{\it Paper organization:} Sec.~\ref{sec:LitReview} discusses prior work; Sec.~\ref{sec:GANs} gives a brief background of GANs; Sec.~\ref{sec:Approach} formulates our approach; Sec.~\ref{sec:InferenceLearning} specifies our inference and learning; Sec.~\ref{sec:Experiments} presents our experiments;  Sec.~\ref{sec:Conclusion} discusses our conclusion.
\vspace{-10pt}
\section{Related Work}\label{sec:LitReview}
\noindent{\bf RBMs} \cite{Hinton_NC2006}, a non-linear generative models. There are multiple time-series variants of RBMs, such as Conditional RBMs \cite{Taylor_JMLR2011}, Temporal RBMs \cite{Sutskever_AISTATS2007}, and Recurrent Temporal RBMs \cite{Sutskever_NIPS2008}, that have been successfully used for human motion generation. Limitations of these models lies in its ability to efficiently handle long-term relationships. Conditional RBMs used auto-regressive nodes to model short-term relationships, Temporal RBMs included directed lateral edge between hidden nodes, which made the model intractable, and finally Recurrent Temporal RBMs which introduced recurrent cells, which had exact inference and learning, but the model complexity prevented it from scaling to more complex and larger number of activities. RBMs have the advantage of learning a probabilistic distribution, but it comes with a price of complex approximate inference and learning algorithms such as contrastive divergence \cite{CD} and variational methods \cite{Salakhutdinov_ICML2009}.

\noindent{\bf RNNs} \cite{Hochreiter1997}, have been leading the way in human motion generation. Some of the problems in the RBMs family are partially addressed using RNNs and Long Short-Term Memory (LSTM). Their success is to be attributed to the possibility to use back-propagation through time \cite{BPTT} and appling stochastic gradient descent \cite{KingmaB14}. The synthesis of realistic human motion, specifically motion completion using RNNs, has been the focus of the community for the past couple of years. Starting with \cite{FragkiadakiLM15} which provided an RNN framework for motion completion and standardized the evaluation metrics for the motion completion task and benchmarked it on Human 3.6M dataset \cite{h36m_pami}. Their approach consisted of an Encoder-Recurrent-Decoder extending the traditional LSTM with an encoder-decoder networks. Structural-RNN \cite{JainZSS16} extends regular RNN approach in \cite{FragkiadakiLM15} and composes RNNs with spatio-temporal graphs. Similarly, \cite{GhoshSAH17} proposed a Dropout Auto-encoder LSTM formulation which enhances the RNN formulation with a de-noising auto-encoder with dropout layers which improves the stability of LSTMs. Recently, \cite{MartinezBR17} modeled velocities, rather than joints angles, using a sequence-to-sequence model with residual architecture and Gated Recurrent Units similar to that used in machine translation. All these methods fail to generate sequences longer than one seconds and the model then drifts. Most recently, \cite{LiZXHL18} proposed auto-conditioned RNNs which links the network’s own predicted output into its future input streams during training. This seems like a simple modification, however, it addresses the problem encountered by prior work, which is that the model is trained on ground truth data, but at inference its making predictions based on noisy data generated by the RNN itself. 

\noindent{\bf GANs} were recently used in language generation \cite{PressBBBW17} formulated using an RNN. With that advancement, training an RNN using a GAN would alleviate the RNN issues such as its susceptibility to noise and drifting.  GANs has been recently used in two non peer-reviewed papers for motion generation, Human Pose-GANs \cite{BarsoumBKL17} and Text2Action \cite{AhnHCYO17}. It seems to have successfully learned temporal representations using a GAN. Both \cite{BarsoumBKL17,AhnHCYO17}\footnote{Both approaches did not report quantitative results and no code available. Without quantitative results it would be hard to judge the models's success or compare aginst.} extended the work of \cite{MartinezBR17} by formulating the a sequence-to-sequence model using a GAN framework. In \cite{AhnHCYO17}, the goal was to synthesize human motion from text, not motion completion. Finally, \cite{KomuraTHHJY17} used Variational Auto-Encoders (VAEs), similar to GANs, VAEs are generative models that learn an explicit density functions. They extended the LSTM approaches above by using a VAE framework consisting of a convolutional encoder and a recurrent decoder. VAEs-LSTM was able to address some of the drift, noise, and stability issues encountered using only RNNs as in \cite{FragkiadakiLM15,JainZSS16,GhoshSAH17,MartinezBR17}.

\noindent{\bf Our Approach} benefits from the findings of prior work. The most related work to ours is \cite{AhnHCYO17,BarsoumBKL17,KomuraTHHJY17,LiZXHL18}. We model human motion using a Wasserstein GAN with Gradient Penalty \cite{GulrajaniAADC17}. Our model conditions on text rather than treating it as a form of supervision. We vary our discriminator and generator between convolutional or recurrent networks to suit the task. We inject noise at different levels of the discriminator rather than just the input. We inject noise at the input by perturbing the input sequence by varying the starting point of the sequence. 
\section{Generative Adversarial Networks}\label{sec:GANs}
\noindent{\bf GANs} \cite{Goodfellow14} consists of two differentiable networks, a generator $G$ and discriminator $D$. The generator's (counterfeiter) objective is to generate data that look real. The discriminator's (appraiser) objective is to assign high value to real data and low value to fake videos. The objective of the game is the average value of real videos subtract the average value of the fake videos. The goal is to learn the parameters of the a generator $\theta_g$ that is able to represent the data distribution, $p_r$, thus the optimization goal is to improve the generator's distribution $p_g$ to match the data distribution $p_r$. The generator network produces a sample $\bar{\mathbf{x}} \sim p_g$, from a latent code $\mathbf{z} \sim p(z)$, where, sampled from a normal distribution $p(z)$, where $\bar{\mathbf{x}} = G(\mathbf{z};\theta_g)$, to be as good as a sample from the data $\mathbf{x}\sim p_r$. The discriminator network is trained to identify whether the sample is generated by the model or a training sample, which requires no supervision. Both networks are trained using the same gradients since they are differentiable. The distance between the distributions is measured using Jensen-Shannon divergence. The formulation solves a dual problem by finding equilibrium in a min-max game with the value function $J(D,G)$ defined in (\ref{eqn:GAN}). 
\begin{equation}
	J(D,G) = \underbrace{\mathbb{E}_{\mathbf{x} \sim p_r}\big[\log D(\mathbf{x})\big]}_{\text{Real}}-\\
	\underbrace{\mathbb{E}_{\mathbf{\bar{\mathbf{x}}} \sim p_g}\big[\log D(\bar{\mathbf{x}})\big]}_{\text{Generated}}
	\label{eqn:GAN}
\end{equation} 
\noindent{\bf Conditional GANs} (CGANs) \cite{GauthierJ14} was introduced to enable conditioning on a class label $\mathbf{y}$ or text with distribution $p_l$. CGANs can be constructed by simply conditioning both the generator and discriminator on $\mathbf{y}$ as defined in (\ref{eqn:CGAN}).
\begin{equation}
	J_{C}(D,G) = \underbrace{\mathbb{E}_{\mathbf{x} \sim p_r,\mathbf{y}\sim p_l}\big[\log D(\mathbf{x},\mathbf{y})\big]}_{\text{Real}}-\\
	\underbrace{\mathbb{E}_{\bar{\mathbf{x}} \sim p_g, \mathbf{y} \sim p_l}\big[\log D(\bar{\mathbf{x}},\mathbf{y})\big]}_{\text{Generated}}
	\label{eqn:CGAN}
\end{equation} 
Even though GAN has a nice theoretical formulation, its optimization is unstable and requires careful balancing between learning the generator and discriminator.

\noindent{\bf Wasserstein GANs} (WGANs) \cite{Arjovsky17} was introduced to improve the stability of GANs. WGANs minimizes the earth mover’s distance between $p_r$ and $p_g$. The formulation restricts the discriminator, $D_w$, to be lipschitz-1 and enforces it by weights clipping. This prevents the discriminator from multiply its valuation by a large constant to make the objective larger. 

\noindent{\bf WGANs with Gradient Penalty} (WGANs-GP) \cite{GulrajaniAADC17} improved WGAN's stability even further by replacing weights clipping with a gradient penalty to the loss function as shown in (\ref{eqn:GPGAN}), where $\mathbf{\hat{x}}\sim p_\mathbf{\hat{x}}$, where $p_\mathbf{\hat{x}}$ is sampled uniformly along straight lines between pairs of points sampled from the data distribution $p_r$ and the generator distribution $p_g$. 
\begin{equation}
\begin{array}{c}
	J_{GP}(D,G) =
	\underbrace{\mathbb{E}_{\mathbf{x}\sim p_d,\mathbf{y}\sim p_l}\big[\log D_w(\mathbf{x},\mathbf{y})\big]}_{\text{Real}}-\underbrace{\mathbb{E}_{\bar{\mathbf{x}} \sim p_g, \mathbf{y} \sim p_l}\big[\log D_w(\bar{\mathbf{x}},\mathbf{y})\big]}_{\text{Generated}}+\\
	\underbrace{\lambda \mathbb{E}_{\mathbf{\hat{x}}\sim p_{\hat{x}}, \mathbf{y} \sim p_l}\big[(\lVert \nabla_{\hat{\mathbf{x}}}D_w(\mathbf{\hat{x}},\mathbf{y}) \rVert_2 -1 )^2\big]}_{\text{Gradient Penalty}}
\end{array}
	\label{eqn:GPGAN}
\end{equation} 
\begin{figure*}
\centering
\includegraphics[width=0.9\textwidth]{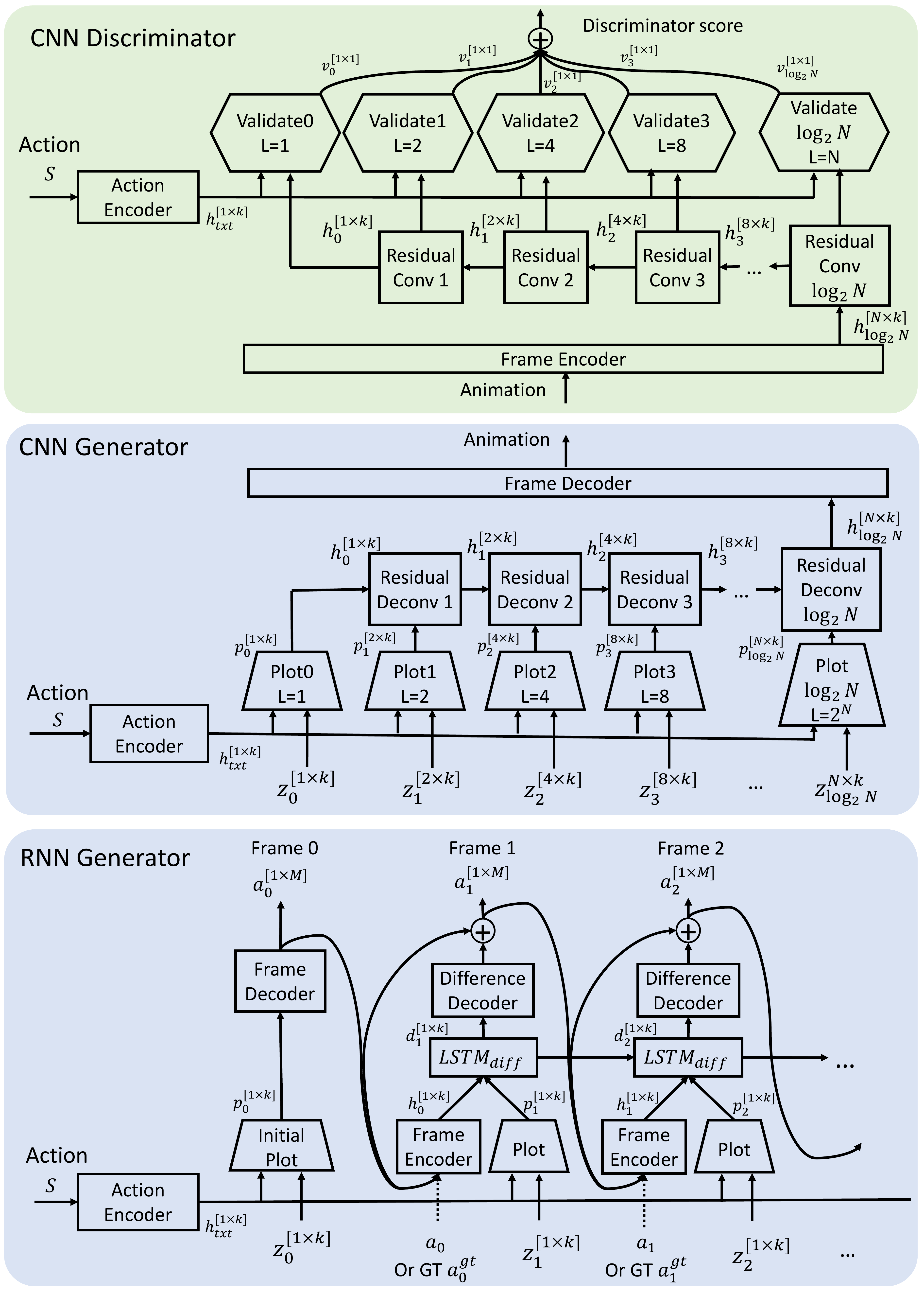}
\caption{The architecture of our DVGANs. The generator network $G$ generates an animations hierarchically by interpolating higher level hidden representations into lower level representations conditioned on textual descriptions. We formulate our approach using two different generators, one using RNN (bottom) and one using CNN (middle). The discriminator network $D$ appraises videos as generated or real through summarizing video into hidden representations at different levels and then validate whether those representations to match the textual descriptions. We formulate  discriminator using a CNN (top). We inject noise at the input of the discriminator by changing the starting point of different actions and do dense validation at the different time resolutions.}
\label{fig:model}
\end{figure*}
\section{DVGANs}\label{sec:Approach}
We propose a new model that combines CNNs and RNNs as discriminator and generator for motion generation using WGAN-GP. Given an action in a sentence format $S$, our goal is to generate an animation by generating a sequence of joint locations and angles. We generate fixed-length animations $A=\{a_1, a_2, ..., a_m\}$ of $N$ frames by $M$ degrees of freedom, at a sampling frequency of $f$ Hz.
\subsection{DVGAN Generators}\label{sec:generator}
We use of two types of generators. The RNN generator is capable of motion completion and generation and the CNN is capable of motion generation.

\paracompact{RNN generator,} $A = G_{RNN}(z,S)$, is first initialized by generating an initial frame using an Multi-Layered Perceptron (MLP), or use initial seed frame(s), followed by generating frame differences using an LSTM-RNN which takes input a concatenation of the previous frame and an animation ``plot'' for the current frame encoded as vectors, and generates the next frame through adding a difference to the previous frame. We define the RNN generator in (\ref{eq:RNNGenerator}) and in Fig.~\ref{fig:model}(bottom).
\begin{equation}
\footnotesize
\begin{array}{rll}
z_t &\sim & \mathcal{N}(0,1),\quad\forall t=0,1,\dots N \\
h_{\text{txt}} &=& \text{LanguageModel}(S) \\
p_0&=& \text{InitialPlot} (h_{\text{txt}},z_{0}) \triangleq \text{Linear} \big\{ \text{ReLU} \lbrack \text{Linear} (z_{0} ) + h_{\text{txt}} \rbrack \big\} \\
a_0&=& \text{Decode} (p_0) \triangleq \text{Linear} (p_0) \\
p_i&=& \text{Plot} (h_{\text{txt}},z_{t}) \triangleq \text{Linear} \big\{\text{ReLU} \lbrack \text{Linear} (z_{t} ) + h_{\text{txt}} \rbrack \big\},\quad\forall t=1,2,\dots N  \\
d_t, \text{state}_t &=& \text{LSTM}_{\text{diff}} \big\{ \lbrack \text{Linear}(a_{t-1}),p_t\rbrack,\text{state}_{t-1} \big\},\quad\forall t=1,2,\dots N \\
a_t &=& a_{t-1}+ \text{Linear}(d_t),\quad \forall t=1,2,\dots N \\
A &=& \{ a_0, a_1, \dots a_N \}
\end{array}
\label{eq:RNNGenerator}
\end{equation}
\paracompact{CNN generator}, $A = G_{CNN}(z,S)$, inspired by the work on super resolution \cite{LedigTHCAATWS17}, the CNN generator starts from a low frame-rate animation, then progressively increase the temporal resolution by upsampling and refining low-frame rate animations into high frame rate ones using a stack of residual deconvolution operations. Each residual convolution module doubles the frame-rate. Generating an animation with $N$ frames would require $\log_2(N)$ residual convolution modules and $\log_2(N)+1$ plot modules. At level $i$, a plot module decides what new motion should be introduced to all frames based on the text representation $h_{txt}$ and $2^i$ frames by $k$ channels gaussian latent code matrix $z_i$. Finally, a 1D convolution is used to decode the hidden representation at the final resolution and output the skeleton animation. We define the CNN generator in (\ref{eq:CNNGenerator}) and in Fig.~\ref{fig:model}(middle).
\begin{equation}
\footnotesize
\begin{array}{rll}
z_i &\sim & \mathcal{N}(0,1), \quad\forall i=0,1,\dots \log_2(N) \\
h_{\text{txt}} &=& \text{LanguageModel}(S) \\
p_i &=& \text{Plot}_i (h_{\text{txt}},z_{i})\triangleq\text{Conv1d} \big\{ \text{ReLU} \lbrack \text{Conv1d} (z_{i} ) + h_{\text{txt}} \rbrack \big\},\forall i=0,1,\dots \log_2(N) \\
h_0 &=& p_0 \\
h_i &=& \text{Residual}_i (h_{i-1},h_{\text{txt}}), \quad\forall i=1,2,\dots \log_2(N)\\
&\triangleq & \text{Upsample1d}(h_{i-1})+\text{Conv1d} \Big( \text{ReLU} \big\{ \text{Upsample1d} \lbrack \text{Conv1d} (h_{i-1}) \rbrack + p_i \big\} \Big)\\
A &=& \text{Decode}(h_{\log_2 (N)}) \triangleq \text{Conv1d} (h_{\log_2 (N)}) 
\end{array}
\label{eq:CNNGenerator}
\end{equation} 
\paracompact{Final cut.} Imagine that we are asked to draw a fixed-length segment of ``walking'' animation. Even before any walking dynamics come into play, we first need to make an important decision: which leg should be in the front in the first frame, or both legs are in the same place because the person hasn't started walking. There is no way of making such decision explicitly when generating periodic motion because where the animation starts has little effect on the validity of the animation. Analogous to common practice in movie-making, we can first produce a longer version of the animation, and then make short cuts out of the longer animation to sample an animation segment. Our CNN generator first runs for one additional convolution layer and generate an animation ``tape'' that has length $2N$, then the final cut layer selects a consecutive length-$N$ segment as the final output under a uniform distribution of cut location. Similarly, the RNN generator runs for $2N$ iterations, and then applies final cut to generate a length $N$ segment. In this way, the top-layer latent variable $z_0$ does not need to explicitly encode information for generating the first frame for periodic motion. 
\subsection{DVGAN Discriminator}\label{sec:discriminator}
 GAN discriminators are very different from classifiers. Rather than predicting the action or object, the job of a GAN discriminator is to tell whether the videos or images are sampled from a real or generated distribution. The discriminator gradients are used to train the network. In order to improve the gradients, recent work \cite{HjelmJCTC18} introduced importance weights to the generated samples to provide a policy gradient which enable training on continuous data without conditioning. Motivated by their work, we propose important modifications to the discriminator: dense validation and data-augmentation.

\paracompact{Dense Validation.} Compositionality in neural networks follows the intuition that high-level representations are formed by combining lower-level representations. For example, to detect an object, a CNN starts by detecting edges, combining them into more complex structures, such as parts in images, and then reasoning is done at the high-level representation. Using the same analogy and applying to human motion, a CNN reasons about high-level motions by combining lower-level motion. The usual a GAN framework, the discriminators goal is to capture the networks' artifacts at the high-level to detect that its fake. But artifacts do not compose like parts do: a higher-level image or motion artifact is not composed of lower-level image artifacts.  For example, a cat might be detected through a combination of eye, nose, mouth and ears, a fake cat is not just a combination of two fake eyes, a fake nose, a fake mouth and two fake ears. It would only takes one fake part for the full image to be fake. Better yet, it could be a combination of real parts in a fake order. Our key observation is that artifacts do not compose like parts compose to objects. Higher-level artifacts such as mis-placements of parts are in fact quite independent of lower-level artifacts. The same observation also applies to RNNs, that artifacts in different frames in a video are more independent than compositional. 

We argue that GAN discriminators need to produce scores at every time-resolution and every frame. Figure.~\ref{fig:model}, we still use a CNN discriminator to encode the content of the input animations to provide context. We add validation modules that inspect intermediate representations at every time-resolution and every frame to ensure that artifacts can be reported immediately. Parameters of validation modules at the same resolution are shared, while parameters of validation modules at different resolutions are independent. Our validation module is a two-layer MLP (implemented as size-1 convolutions for parameter sharing over frames) that matches a motion representation at a particular time-resolution and frame, with the action sentence representation. The discriminator produces a final score as a convex combination of validation scores at different resolutions (the weights are parameters of the discriminator network). This is useful for WGAN where the discriminator network as a whole is restricted to Lipschitz-1 because doing so enable validation modules at different resolutions to adjust their local Lipschitz constants through adjusting the weights. We find this design to greatly improve the generation quality compared to scoring only the final convolution layer or every other convolution layer.

\paracompact{Data augmentation.} We treat that artifact detection as a translation-invariant problem. We randomly purturb the input video to the discriminator by $\lbrack -\frac{N}{2},\frac{N}{2}\rbrack$ and zero-pad the unknowns. That improves the translation-invariance aspect of the discriminator. It also reduces one typical failure pattern that the motion pattern of first half of the generated video is different from the second half. We hypothesize that this limitation is due to ability of the discriminator to detect higher-level artifacts using low sampling resolutions at top layers due to downsampling. The discriminator fails to detect the difference between motion patterns in the first half and the second half of the animation. Stochastically shifting the input videos perturbs the sampling grid for higher-level layers, enabling the detection of such failure patterns.

\paracompact{CNN Discriminator}\footnote{We explored using RNN discriminator, which is similar to that of the RNN generator. The optimization did not converge with WGAN-GP due to dissipating gradients. More discussions are available in supplementary material.}, $y = D_{CNN}(A,S)$, mirrors the CNN generator. The input video is first encoded into a $N$ frames by $k$ channels representation through size-1 1D convolution. A residual convolution network. Incorporating our dense validation design, we validate the hidden representations at every resolution and every frame using $\log_2 (N)$ validation modules. For generators with final cut, that would require $\log_2(N)+1$ residual convolution modules and $\log_2(N)+2$ plot modules. Finally, the validation scores are combined through a learned convex combination. We define the CNN discriminator in (\ref{eq:CNNDiscriminator}) and illustrate it in Fig.~\ref{fig:model}(top).
\begin{equation}
\footnotesize
\begin{array}{rll}
h_{\text{txt}} &=& \text{LanguageModel}(S) \\
h_{\log_2 (N)} &=& \text{Encode}(A) \triangleq \text{Conv1d} (A) \\
h_i &=& \text{Residual}_i (h_{i+1}),\quad\forall i=0,1,\dots \log_2(N)-1\\
&\triangleq & \text{Dnsample1d}(h_{i+1})+\text{Conv1d} \big( \text{ReLU} \big\{ \text{Dnsample1d} \lbrack \text{Conv1d} (h_{i+1}) \rbrack \big\} \big) \\
s_i &=& \text{Validate} (h_i) \quad\forall i=0,1,\dots \log_2(N)\\
&\triangleq & \text{Mean} \Big( \text{Conv1d} \big\{ \text{ReLU} \lbrack \text{Conv1d} (h_i) + \text{Conv1d} (h_{txt}) \rbrack \big\} \Big)\\
y   &=&  \sum_{i}{ e^{w_i} s_i}
\end{array}
\label{eq:CNNDiscriminator}
\end{equation}

\section{Inference and Learning}\label{sec:InferenceLearning}

\subsection{Inference}
We address the following two tasks, motion completion and generation.

\paracompact{Motion Generation} is not a popular of a task. None of the prior work focused on motion generation from scratch with exception to \cite{AhnHCYO17}. We are the first to quantitatively evaluate this task. Both RNN-DVGAN and CNN-DVGAN are capable of motion generation since we do not require conditioning on seed frames to generate an action. We do a forward pass to generate the animations using our CNN generator $A = G_{CNN}(z,S)$ or our RNN generator $A = G_{RNN}(z,S)$.


\paracompact{Motion Completion} is a very popular task in the human motion modeling community~\cite{Taylor_JMLR2011,FragkiadakiLM15,GhoshSAH17,MartinezBR17}. Given a sequence of $n$ initial ground truth frames $\{a^{gt}_0,a^{gt}_1,\dots,a^{gt}_n\}$, the goal is to predict frames from $n+1$ to $N$ $\{a_{n+1},a_{n+2},\dots,a_N\}$. We set $a_t=a^{gt}_t, t=1,2,\dots,n$ as input to our RNN-DVGAN and our model generates $a_{n+1}\sim a_N$ conditioned on not only text but also ground truth frames. This can be seen as learning the motion predictor using a conditional-GAN loss instead of the commonly-used L2 loss in motion prediction. Learning to combine the previous frame with a noisy animation ``plot'' to predict the next frame is an extension to ~\cite{LiZXHL18} which observes better motion completion from explicitly perturbing the previous frame as the input to predict the next frame. 

\subsection{Learning}
We optimize the WGAN-GP loss in (\ref{eqn:GPGAN}) to train our generators. The generators are not restricted to predicting the ground truth motion, but can generalize over different motions and generate multiple modes of realistic motions for the given action without being penalized. As a result, our generators shows substantially improved robustness when generating long sequences. As a result, our DVGAN trained on 5 seconds video clips is capable of generating realistic motions reaching 300 seconds and beyond.

We first preprocess training data to represent body-joint angles in exponential maps, and then normalize the skeleton representation by subtracting mean and dividing by standard deviation. For both generators and discriminator, we set the hidden size for all modules to be $k=256$. Our language model for action description encoding is a 2-layer LSTM with 256 hidden units, and the word embeddings are learned from scratch. Our $LSTM_{diff}$ module in the RNN generator is also a 2-layer LSTM with 256 hidden units. 

For optimization we pair either a CNN or a RNN generator with the CNN discriminator, and learn parameters using WGAN-GP (\ref{eqn:GPGAN}) with $\lambda=10$, where the discriminator maximizes the loss and the generator minimizes the loss. We run the Adam optimizer \cite{KingmaB14} with learning rate $1\times10^{-4}$ for 20,000 iterations. In each iteration D is updated for 10 times before G is updated once. All the parameter tuning is based on a grid search to find the optimal set.
\section{Experiments}\label{sec:Experiments}
\subsection{Datasets}
\paracompact{H3.6M~\cite{h36m_pami}} is a widely used dataset for benchmarking human motion prediction~\cite{Taylor_JMLR2011,FragkiadakiLM15,GhoshSAH17,MartinezBR17}. The H3.6M dataset contains 7 actors performing 15 different activities including eating, discussing, giving directions, greeting, phoning, posing, purchasing, smoking, sitting, sitting down, taking photos, waiting, walking, walking dog and walking together, each in two trials captured at 50Hz. We adopt the same preprocessing procedure from~\cite{JainZSS16,MartinezBR17} and represent body-joint angles in exponential maps. There are 32 body joints for each body skeleton hierarchy. The length of the videos are between 20 seconds and 130 seconds. Overall, the H3.6M dataset contains 15 actions, 210 videos totaling 3 hours. Following previous works~\cite{Taylor_JMLR2011,FragkiadakiLM15,GhoshSAH17,MartinezBR17}, we reserve sequences from actor 5 for testing and use the sequences of all other actors for training. Note that this split is not only for motion completion but also for generation and retrieval experiments. 

\paracompact{CMU Mocap~\cite{CMUMocap}} is a large-scale motion capture dataset of open-ended actions. It contains 2,548 videos from 113 actors performing 1,095 unique activities captured at 120Hz. There are activities with different styles and transitions such as ``walk on uneven terrain'', ``dance - expressive arms, pirouette'', ``punch and kick'' and ``run to sneak''. Such fine-grained activities are very useful for studying action generation given description. \cite{ButepageBKK17} evaluated their approach on a part of the dataset for motion completion, however, it was not standardized to compare against. \cite{Taylor_JMLR2011} used a set of different walking and running sequences of the dataset and evaluated it qualitatively. \cite{LiZXHL18} also used 4 classes, variation of dancing and walking. We focus on actions that are at least 8 seconds long which narrows the dataset down to 1,125 videos. We randomly partition the dataset, using 757 sequences for training and 368 sequences for testing. We use BVH format~\cite{CMUMocapBVH} where the skeletons are normalized to 31 joints. Similar to H3.6M, we pre-process the joint angles into the exponential map representation. Our train+test splits of the CMU Mocap dataset contain 573 actions, 1,125 videos totaling 8 hours.


\subsection{Evaluation Metrics}
Evaluation of generative models has been an active area of research~\cite{BarrattS18}. Since our interest is in motion generation from scratch, with focus on the diversity and realism of the samples, we adopt the inception score~\cite{SalimansFZCRX16}. It measures the quality of unconditioned generation and the accuracy of action retrieval using the generated videos. This measures how correlated the generated videos are to the action descriptions which they are conditioned on.
%
%

\paracompact{Animation Retrieval Accuracy}, 
%
is formulated as follows: given a fixed-length animation clip, the goal is to retrieve its original description from a pool of $K$ descriptions. We use recall as our retrieval evaluation metric for the top 1, 3, 5 and 10 predictions. This measures how often is the ground truth description in the top 1, 3, 5 and 10 high-scoring descriptions respectively. Inspired by image-caption ranking~\cite{Karpathy_2015,Kiros_2015}, we learn a ranker network that encodes the action description into a vector, encodes the animation into a vector and then predicts their dot product as a matching score between an action description and an animation. For action description, we use a LSTM language model with 2 layers of 1024 units as the encoder. For the animation, we experiment with both using a CNN or a 2-layer LSTM RNN with hidden size $k=1024$ to encode the animation clip\footnote{Due to space limits details about the animation encoder are in supplement material.}. We learn the ranker through minimizing the negative log-likelihood (NLL) loss of retrieving the correct description out of $K$ descriptions~\cite{Lin_2016}, using Adam optimizer with learning rate $1\times10^{-4}$, over 100 epochs. For the CMU Mocap dataset, we set $K=250$. For H3.6M, we set $K=15$ to cover all the 15 actions.
\begin{table*}[!htbp]
\scriptsize
\centering
\begin{tabular}{ l |c c c c | c c c c }
\toprule
\multicolumn{9}{c}{Action Retrieval Performance}\\
\midrule
Approach & \multicolumn{4}{c}{CMU Mocap} & \multicolumn{4}{c}{H3.6M} \\
 & R@1 & R@3 & R@5 & R@10& R@1 & R@3 & R@5 & R@10 \\
\midrule
Random  	&  0.4 	& 1.2	& 2.0  	& 4.0 &  6.7 	& 20.0 & 33.3  & 66.7\\
CNN ranker  &  36.5 & 45.6 	& 49.1  & 53.8 &  44.5 & 69.4 & 82.6  & 96.4 \\
RNN ranker  &  \textbf{39.7} & \textbf{48.3} 	& \textbf{51.4}  & \textbf{58.7} &  \textbf{47.2} & \textbf{73.9} & \textbf{85.2}  & \textbf{97.0} \\
\bottomrule
\end{tabular}
\caption{Animation retrieval accuracy on CMU Mocap: $N=32$, $f=4\text{Hz}$ and H3.6M: $N=64$, $f=12.5\text{Hz}$. On both datasets, RNN and CNN ranking models perform well. We proceed to using the RNN ranker to evaluate animations generated from GAN.}
\label{table:retrieval}
\end{table*}
For CMU mocap, we train models for clip with length $N=32$ at $f=4\text{Hz}$. For H3.6M, the clips are length $N=64$ at $f=12.5\text{Hz}$. Table~\ref{table:retrieval} shows the retrieval accuracy of our models on the test split. On CMU Mocap dataset our RNN model achieves a retrieval accuracy of 39.7\% top-1 and 48.3\% top-3. On H3.6M dataset, our RNN model achieves a retrieval accuracy of 47.4\% top-1 and 72.8\% top-3. The accuracies are high considering the input is a clip rather than a whole video. For example a ``walk and run'' action might contain both a clip of running and a clip of walking, which might be matched to ``walk'' and ``run'' respectively rather than the combined action ``walk and run''. 

For GAN evaluation, for a animation generated from a test set action description, we use the RNN model to compute the action retrieval accuracy at top 1, 3, 5, 10 predictions of retrieving the action description from a set of $K=250$ and $K=15$ test set action descriptions, for CMU mocap and H3.6M respectively.
%



\paracompact{Inception Score}, was first proposed to evaluate image generated from GANs. Intuitively, it encourages each generated image to be specific to very few object classes and diverse among all generated images. For image generation inception score has been shown to correlate well with human perception~\cite{SalimansFZCRX16}. The higher the inception score is, the better the quality is of the generated image.  
%
To compute the inception score, we first run our RNN ranker to compute matching score between the input animation and each of the top-$K$ most popular action descriptions in the test set. We use softmax of the score as the distribution $p(\text{action}\;|\;\text{animation})$, and then inception score is defined in (\ref{eqn:Inception}). We use $K=250$ and $K=15$ descriptions for CMU Mocap and H3.6M respectively.
\begin{equation}
\footnotesize
s_{\text{inception}} \triangleq \sum_{\text{action}} \text{KL}\big(p(\text{action}\;|\;\text{animation})\;||\;p(\text{action})\big) = \mathbb{H}(\text{action})-\mathbb{H}(\text{action}\;|\;\text{animation})
\label{eqn:Inception}
\end{equation}
%
%
\begin{figure*}[t]
\centering
\fbox{\includegraphics[width=\textwidth]{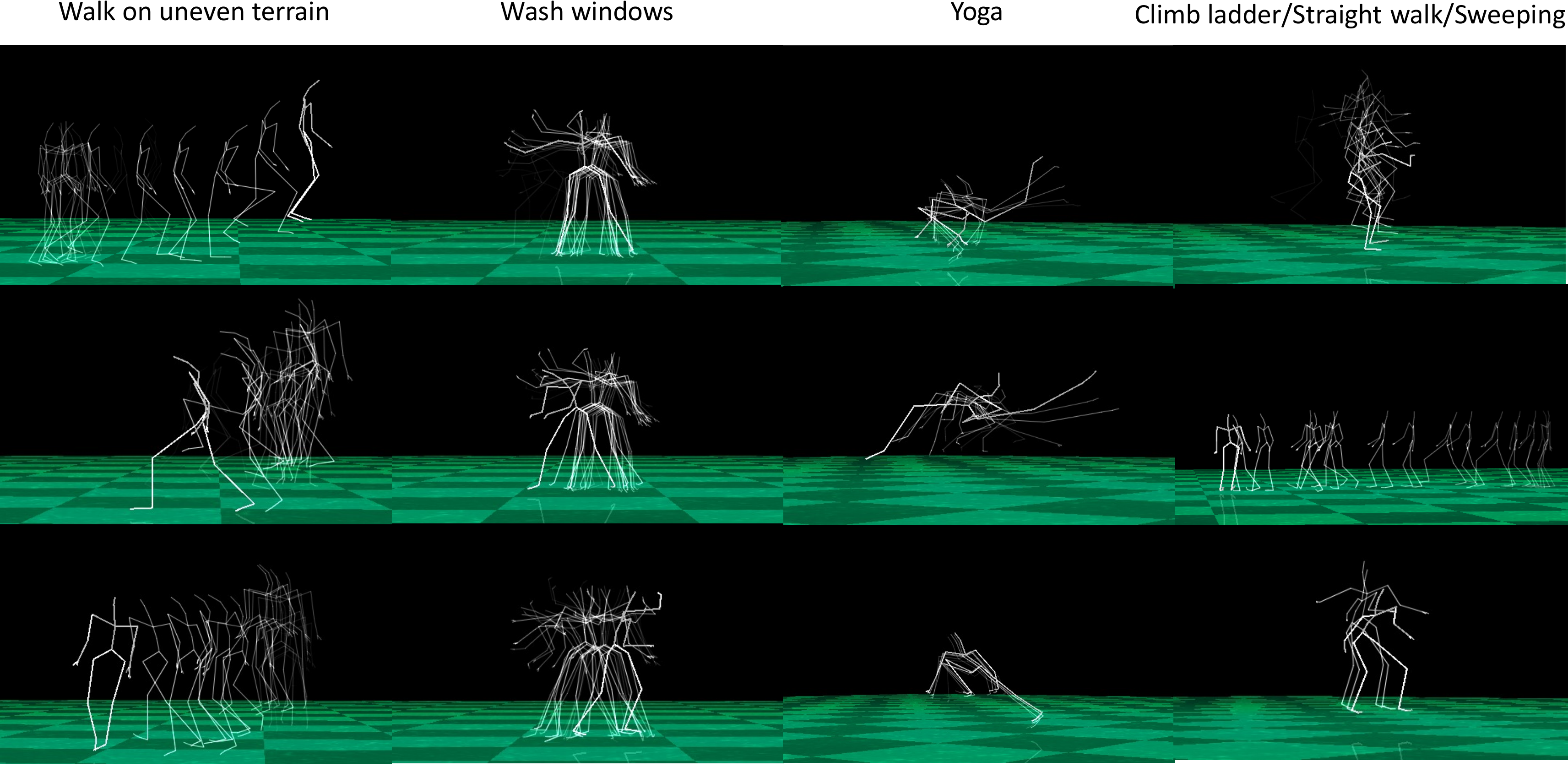}}\vfill%
\vspace{2pt}
\fbox{\includegraphics[width=\textwidth]{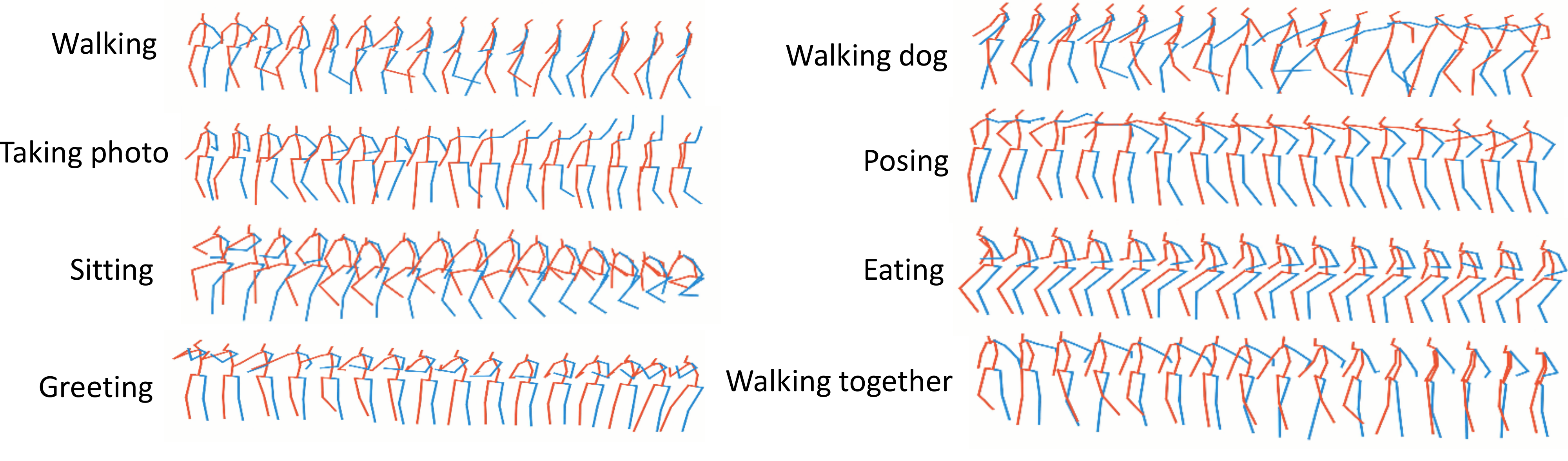}}
\caption{Qualitative results on CMU Mocap dataset (top) and H3.6M (bottom). As you can see, DVGANs is able to generate smooth actions diversely on both datasets.}
\label{fig:qualitative}
\end{figure*}
\subsection{Motion Generation}
\paracompact{Qualitative Results.} We evaluate our models generation abilities qualitatively on both datasets.
On the CMU Mocap dataset, we train our CNN-DVGAN to generate 8-second animations with $N=32$ and $f=4\text{Hz}$ shown in Fig.~\ref{fig:qualitative}(top). On CMU Mocap for popular actions in the training set such as ``walk on uneven terrain'', `` wash windows'' and ``yoga'' with more than 100 seconds of video available for training and even less common actions such as ``climb ladder'' with 40 seconds of video for training, the generated videos look realistic and show different variations\footnote{Additional videos are available in the supplement material.}. This shows how the model is able to generalize over all the activities and generate a diverse set of motions. Even though an action such as ``straight walk'' is unseen in the training set, our model is still able to transfer knowledge from various ``walk'' actions and generate realistic motion. Treating action labels as text enables our model to transfer knowledge from familiar actions and generate animation for unseen actions.
On the H3.6M dataset, we train the CNN-DVGAN to generate 5-second animations with $N=64$ and $f=12.5\text{Hz}$ shown in Fig.~\ref{fig:qualitative}(bottom).  All 15 actions have abundant training data. Our CNN-DVGAN model is able to generate diverse and realistic actions.

\paracompact{Quantitative Analysis.} We conduct ablation studies to show the effect of each improvement proposed in sec.~\ref{sec:Approach}. We quantitatively study the impact of model architecture (CNN, RNN), dense validation, data augmentation and final cut on CMU Mocap with $N=32$ and $f=4\text{Hz}$, and H3.6M with $N=64$ and $f=12.5\text{Hz}$. We measure the effect of having a validator at every resolution (dense), versus having a validator at the final convolution layer (final), or every other layer (mod 2). For data augmentation and final cut, we conduct experiments with turning those options on or off.
\begin{table*}[t]
\centering
\begin{tabular}{ l |c c c | c |c c c c }
\toprule
\multicolumn{9}{c}{CMU Mocap Ablation Study}\\
\midrule
Generator	&Validation	& Data	& Final & Inception & \multicolumn{4}{c}{Action Retrieval} \\
			&  			& Aug. 	& Cut 	& Score & R@1 	& R@3 	& R@5 	& R@10 	\\
\midrule
CNN		& final	&			&		& 3.62	& 40.6	& 50.4	& 53.1	& 56.5	\\
CNN		& mod 2	&			&		& 3.80	& 47.6	& 56.6	& 58.4	& 60.9	\\
CNN		& dense	&			&		& 3.88	& 51.0	& 58.0	& 59.7	& 61.9	\\
CNN		& dense	&	y		&		& 3.85	& 52.2	& 58.4	& 60.3	& 63.0	\\
CNN		& dense	&	y		&	y	& \textbf{3.89}	& \textbf{52.3}	& 57.6	& 58.8	& 60.7	\\
\midrule
RNN		& dense	&			&		& 3.62	& 50.4	& 58.3	& 60.9	& 63.5	\\
RNN		& dense	&	y		&		& 3.61	& 49.9	& \textbf{59.0}	& \textbf{61.2}	& \textbf{64.2}	\\
RNN		& dense	&	y		&	y	& 3.53	& 47.3	& 56.5	& 59.6	& 62.9	\\
\midrule
\multicolumn{3}{c}{Real video samples}&	 	& 3.74	& 39.7	& 48.3	& 51.4	& 58.7	\\
\bottomrule
\end{tabular}
\begin{tabular}{ l |c c c | c |c c c c }
\toprule
\multicolumn{9}{c}{H3.6M Ablation Study} \\
\midrule
Generator	&Validation	& Data	& Final & Inception & \multicolumn{4}{c}{Action Retrieval} \\
			&  			& Aug. 	& Cut 	& Score & R@1 	& R@3 	& R@5 	& R@10 	\\
\midrule
CNN		& final	&			&		& 2.49		& 91.5		& 97.2		& 98.6		& 99.8	\\
CNN		& mod 2	&			&		& 2.55		& 97.2		& 99.8		& 99.9		& 100.0	\\
CNN		& dense	&			&		& 2.53		& 97.5		& 99.8		& 100.0		& 100.0	\\
CNN		& dense	&	y		&		& 2.56		& 98.6		& 100.0		& 100.0		& 100.0	\\
CNN		& dense	&	y		&	y	& {\bf2.58}	& {\bf98.7}	& {\bf100.0}	& {\bf100.0}& {\bf100.0}	\\
\midrule
RNN		& dense	&			&		& 2.51		& 97.7		& 99.9		& 99.9		& 100.0	\\
RNN		& dense	&	y		&		& 2.50		& 96.5		& 99.7		& 99.9		& 100.0	\\
RNN		& dense	&	y		&	y	& 2.36		& 88.5		& 98.3		& 99.9		& 100.0	\\
\midrule
\multicolumn{3}{c}{Real video samples} & 	& 1.48	& 47.2	& 73.9	& 85.2	& 97.0	\\
\bottomrule
\end{tabular}
\caption{Ablation studies of our model evaluated on CMU mocap and H3.6M.}
\label{table:ablation}
\vspace{-28pt}
\end{table*}

The results are shown in Table~\ref{table:ablation}. It is obvious that the dense validation significantly improved the results for both RNNs and CNNs. The improvements of final cut and data augmentation are marginal, however, they still contribute to the retrieval accuracy. They also had impact on convergence for RNNs. For our two models, both generators performed reasonably well. The CNN generator achieved a higher inception score, which is what we expected. CNNs are well suited for generation from scratch since they generate the full sequence at once, while RNNs are well suited for motion completion task as we will see in the motion completion evaluation\footnote{Our study showcase RNNs and CNNs as viable for text to action generation.}. 

\subsection{Motion Completion}
\begin{figure}[t]
\centering
\includegraphics[width=0.9\textwidth]{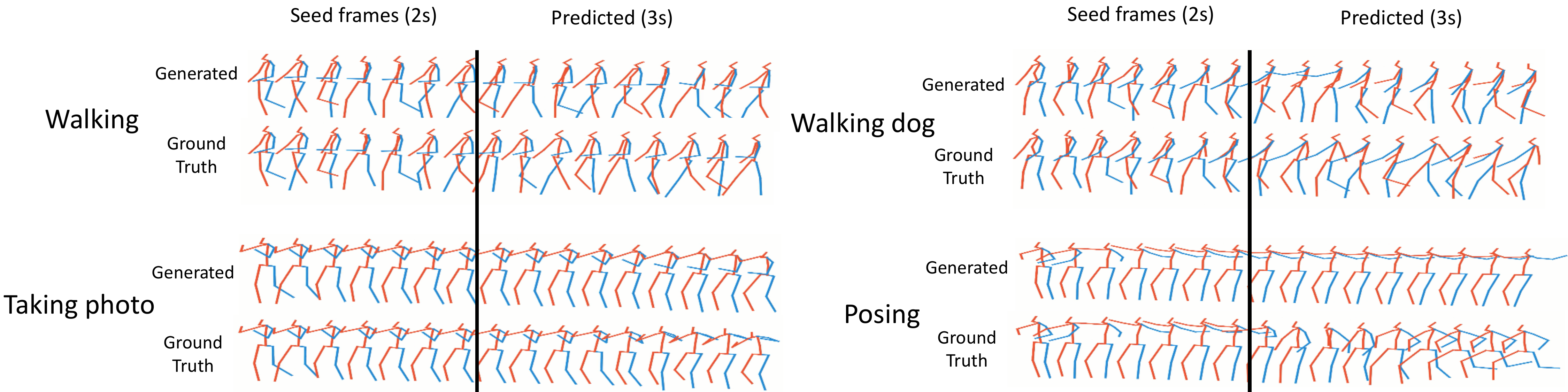}
\caption{Qualitative results for motion completion on H3.6M dataset.}
\label{fig:qual_pred}
\end{figure}
\paracompact{Qualitative Results.} Figure~\ref{fig:qual_pred} shows our qualitative results for motion completion using RNN-DVGAN comparing to ground truth. Our RNN-DVGAN is able to complete sequences with realistic frames. The L2 loss however, does not correlate well with the perceived quality of the generated video\footnote{RNN-DVGAN is trained for predicting only 5 seconds of video, however, it is able to generate realistic ``walking'' videos up to 300 seconds. See supplemental material.}. 

\begin{table*}[!htbp]
\centering
\begin{tabular}{ l |c c c c |c c c c }
\toprule
\multicolumn{9}{c}{\textbf{H3.6M motion prediction error with 2-second initialization}}\\
\midrule
			& \multicolumn{4}{c}{Walking} 	&  \multicolumn{4}{c}{Eating} 		\\
Approach	& 80ms	& 160ms	& 320ms	& 400ms 	&  80ms	& 160ms	& 320ms	& 400ms  \\
\midrule
ERD~\cite{FragkiadakiLM15}				& 0.93	& 1.18	& 1.59	& 1.78	& 1.27	&  1.45	& 1.66	& 1.80 	\\
LSTM-3LR~\cite{FragkiadakiLM15}			& 0.77	& 1.00	& 1.29	& 1.47	& 0.89	& 1.09	& 1.35	& 1.46 	\\
SRNN~\cite{JainZSS16}					& 0.81	& 0.94	& 1.16	& 1.30	& 0.97	& 1.14	& 1.35	& 1.46 	\\
Seq2seq~\cite{MartinezBR17}				& \textbf{0.28}	& \textbf{0.49}	& \textbf{0.72}	& \textbf{0.81}	& \textbf{0.23}	& \textbf{0.39}	& \textbf{0.62}	& \textbf{0.76} 	\\
Zero-velocity baseline							& 0.39	& 0.68	& 0.99	& 1.15	& \underline{0.27}	& \underline{0.48}	& \underline{0.73}	& \underline{0.86} 	\\
\midrule
RNN-DVGAN (Ours)					& \underline{0.36}	& \underline{0.59} 	& \underline{0.89}	& \underline{1.00}	& 0.39	& 0.61 	& 1.01 	& 1.20 	\\
\bottomrule
			& \multicolumn{4}{c}{Smoking} 	&  \multicolumn{4}{c}{Discussion} 		\\
Approach	& 80ms	& 160ms	& 320ms	& 400ms	&  80ms	& 160ms	& 320ms	& 400ms  \\
\midrule
ERD~\cite{FragkiadakiLM15}				& 1.66	& 1.95	& 2.35	& 2.42	& 2.27	&  2.47	& 2.68	& 2.76	\\
LSTM-3LR~\cite{FragkiadakiLM15}			& 1.34	& 1.65	& 2.04	& 2.16	& 1.88	& 2.12	& 2.25	& 2.23 	\\
SRNN~\cite{JainZSS16}					& 1.45	& 1.68	& 1.94	& 2.08	& 1.22	& 1.49	& 1.83	& 1.93 	\\
Seq2seq~\cite{MartinezBR17}				& 0.33	& 0.61	& 1.05 	& 1.15	& \textbf{0.31}	& \underline{0.68}	& \underline{1.01}	& \underline{1.09} 	\\
Zero-velocity baseline							& \textbf{0.26}	& \textbf{0.48}	& \underline{0.97}	& \textbf{0.95}	& \textbf{0.31}	& \textbf{0.67}	& \textbf{0.94}	& \textbf{1.04} 	\\
\midrule
RNN-DVGAN (Ours)		& \underline{0.28}	& \underline{0.49}	& \textbf{0.85}	& \underline{0.97}	& 0.50	& 0.86	& 1.31	& 1.53	\\
\bottomrule
\end{tabular}
\caption{Motion prediction error. The error is measured in euclidean angle error over all body joints. Our RNN-DVGAN performs competitively with existing approaches on short-term motion completion, although our RNN-DVGAN focuses on long-term generation beyond 2 seconds and is not optimized for this short-term motion prediction.}
\label{table:prediction}
\end{table*}

\paracompact{Quantitative Analysis.} We follow \cite{MartinezBR17} evaluation using H3.6M. A model is first initialized with 2 seconds of ground truth, and then predict the future frames in a short time window. The evaluation metric is euclidean differences in euler angles of all joints for different animation length. We use an RNN-DVGAN with $N=64$ and $f=12.5\text{Hz}$ as stated in sec.~\ref{sec:InferenceLearning}. For motion completion we initialize the RNN-DVGAN with 25 ground truth frames. We compare against~\cite{FragkiadakiLM15,JainZSS16,MartinezBR17} and a zero-velocity baseline\footnote{The zero-velocity baseline uses the last initial ground truth frame as its prediction which was shown by \cite{MartinezBR17} as a strong baseline for motion prediction.}. The results for four actions ``walking'', ``eating'', ``smoking'' and ``discussion'' are shown in Table~\ref{table:prediction}. Although our RNN-DVGAN focuses on long-term generation beyond 2 seconds and is not optimized for this short-term motion prediction error, our RNN-DVGAN still performs competitively with existing approaches on short-term motion completion. Note that this metric focuses on short-term evaluations and is not indicative of long term prediction performance. Because actions start to exhibit different modes in the long term.
\section{Conclusion}\label{sec:Conclusion}
We propose a novel DVGANs model for motion generation and completion conditioned on text, which uses dense validation, data augmentation and final cut. We evaluate DVGANs both qualitatively on motion generation and completion. We are the first to report inception scores and action retrieval accuracy and benchmark on the entire CMU Mocap Database for motion generation.

\section*{Acknowledgements}
This work is funded in part by DARPA-Communicating with Computers (W911NF-15-C-0246) and DARPA-Explainable AI (FA8750-17-C-0115). The views, opinions, and/or conclusions contained in this paper are those of the author and should not be interpreted as representing the official views or policies, either expressed or implied of the DARPA or the DoD.
\clearpage
\bibliographystyle{splncs}
\bibliography{egbib}
\clearpage
\beginsupplement
\section{Generated Animations}
\subsection{Motion Generation}
Following the protocol described in Section 6.3, We use our CNN-DVGAN to generate animations for the H3.6M dataset and the CMU Mocap dataset.

For the H3.6M dataset, we randomly sample $N=64$ frames and $f=12.5\text{Hz}$ animations using the CNN-DVGAN. The generated animations are available in mp4 format in \url{/h36m-generation/{action}_vid{id}.mp4}.

For the CMU Mocap dataset, we randomly sample $N=32$ frames and $f=4\text{Hz}$ animations using the CNN-DVGAN. The generated animations are available in bvh\footnote{Can be viewed with Autodesk FBX review \url{https://www.autodesk.com/products/fbx/fbx-review}. For Figure.2 we used bvh visualization software by Chris Mizerak and Jared Sohn, \url{http://research.cs.wisc.edu/graphics/Courses/838-s2002/Students/mizerak/project1.htm}, CS838 Class Project, University of Wisconsin, 2002 (requires additional dlls to run on Windows 10).} format in \url{/mocap-generation/ID{action_id}_{action}_vid{id}.bvh}. We sort the actions by their numbers of frames in the train set. The number of training frames for each action is listed in spreadsheet \url{/mocap-actions.xlsx}.

\subsection{Motion Completion}

For the H3.6M dataset, we perform motion completion using the $N=64$ frames and $f=12.5\text{Hz}$ RNN-DVGAN. We initialize the model with 25 frames (2 seconds), and randomly sample the rest 39 frames. The completed animations are available in mp4 format in \url{/h36m-completion/{action}_{id}_{80ms err}.mp4}. The corresponding ground truth sequences are \url{/h36m-completion/{action}_{id}_gt.mp4}.

\subsection{Long Timespan Motion Generation and Completion}

\begin{figure*}
\centering
\includegraphics[width=0.9\textwidth]{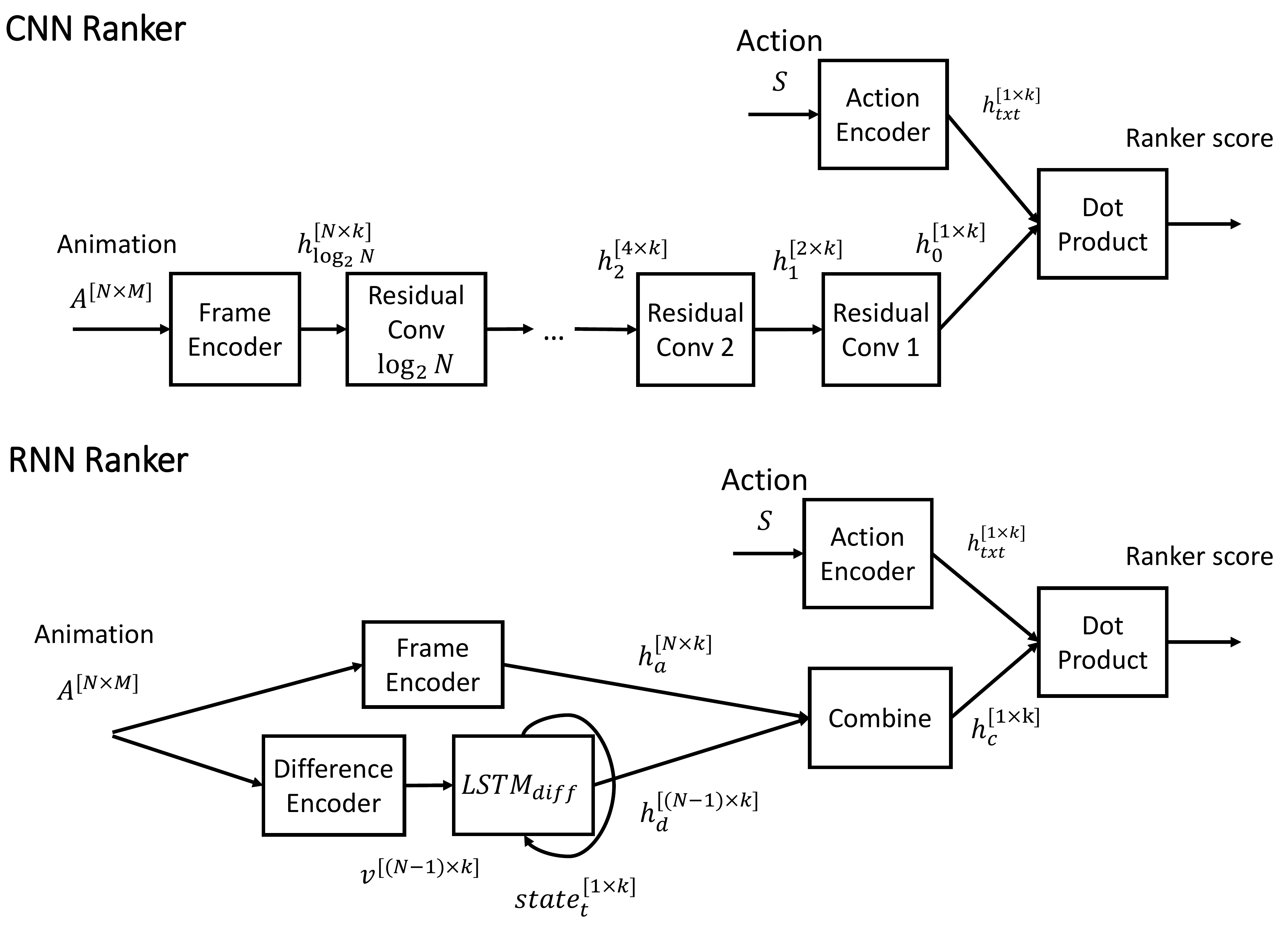}
\caption{The architecture of our CNN and RNN rankers.}
\label{fig:ranker}
\end{figure*}

On H3.6M, although only trained $N=64$ frames, our RNN-DVGAN is capable of evaluating on longer $N$s and generating realistic long animations with $N=1,000$ and $3,000$ frames. That is more than $40\times$ the number of frames it was trained for. 
We generate animations for ``walking'' on the H3.6M dataset, available in \url{/h36m-long/walking_1000.mp4} and \url{/h36m-long/walking_3000.mp4}. Notice how the $N=3,000$ video maintains a walking status over a long period of time. That illustrates the robustness of our approach.

\section{Details about the Ranker Architectures}

As discussed in Section 6.2, Inspired by image-caption ranking~\cite{Karpathy_2015,Kiros_2015}, we learn a ranker network $F(S,A)$ that encodes the action description $S$ into a vector, encodes the animation $A$ into a vector and then predicts their dot product as a matching score between an action description and an animation. For action description, we use a LSTM language model with 2 layers of 1024 units as the encoder. For the animation, we experiment with both using a CNN or a 2-layer LSTM RNN with hidden size $k=1024$ to encode the animation clip.
Figure~\ref{fig:ranker} shows the architecture of both our CNN (top) and RNN (bottom) ranker architectures.

\begin{figure*}
\centering
\includegraphics[width=0.9\textwidth]{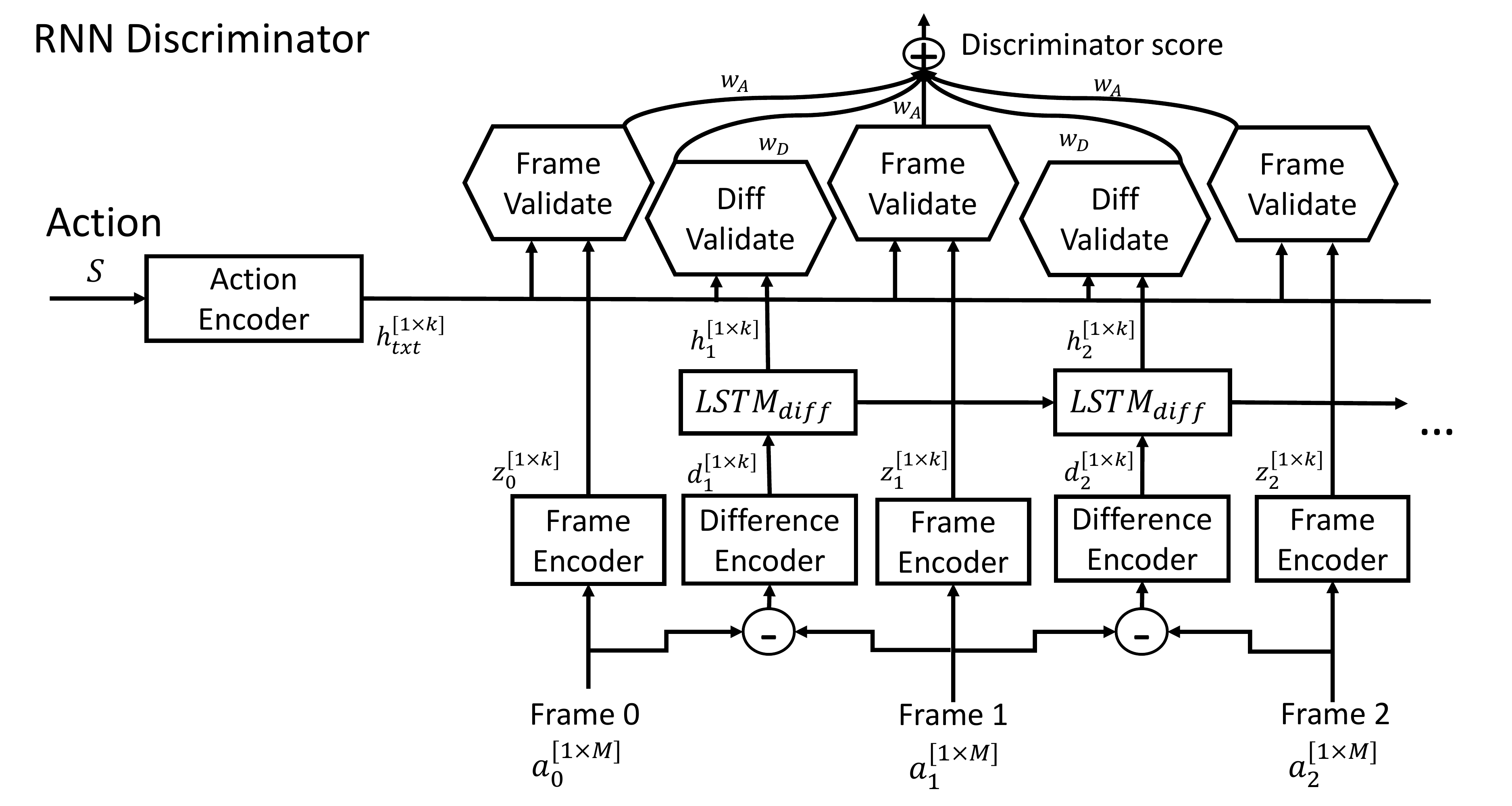}
\caption{The architecture of our RNN discriminator.}
\label{fig:rnn_disc}
\end{figure*}

The CNN ranker (Figure~\ref{fig:ranker} top) encodes the animation through a residual CNN. It is defined as Equation~\ref{eq:CNNRanker}.

\begin{equation}
\footnotesize
\begin{array}{rll}
h_{\text{txt}} &=& \text{LanguageModel}(S) \\
h_{\log_2 (N)} &=& \text{Encode}(A) \triangleq \text{Conv1d} (A) \\
h_i &=& \text{Residual}_i (h_{i+1}),\quad\forall i=0,1,\dots \log_2(N)-1\\
&\triangleq & \text{Dnsample1d}(h_{i+1})+\text{Conv1d} \big( \text{ReLU} \big\{ \text{Dnsample1d} \lbrack \text{Conv1d} (h_{i+1}) \rbrack \big\} \big) \\
F(S,A) & \triangleq &  \langle h_{\text{txt}} , \frac{h_0}{|| h_0 ||_2} \rangle
\end{array}
\label{eq:CNNRanker}
\end{equation}

The RNN ranker (Figure~\ref{fig:ranker} bottom) uses a 2-layer LSTM-RNN to encode the frame differences as a representation of motion dynamics, uses a size-1 1D convolution to encode the frames as a representation of poses, and then combine both representations averaged over all frames through linear projections as the final animation representation. 
It is defined as Equation~\ref{eq:RNNRanker}.

\begin{equation}
\footnotesize
\begin{array}{rll}
h_{\text{txt}} &=& \text{LanguageModel}(S) \\
h_a & = & \text{Encode} (A) \triangleq \text{Conv1d} (A) \\
v & = & \text{DiffEncoder} (A) \triangleq  \text{Conv1d} \lbrack A(2:N,:)-A(1:N-1,:) \rbrack \\
h_d, \text{state}_{N-1} &=& \text{LSTM}_{\text{diff}} ( v,\text{state}_{0} ) \\
h_c &=& \text{Combine} (h_a,h_d) \\
& \triangleq & \text{Linear} \big( \text{Concat} \{ \text{Mean} \lbrack \text{Conv1d}(h_d) \rbrack ,\text{Mean}(h_a) \} \big) \\
F(S,A) & \triangleq &  \langle h_{\text{txt}} , \frac{h_c}{|| h_c ||_2} \rangle
\end{array}
\label{eq:RNNRanker}
\end{equation}

For both the CNN and RNN rankers, we use size $k=1024$ for the hidden states.

\section{On an RNN Discriminator}

We mirror the RNN generator in Section 4.1 to build a RNN discriminator $y=D_{\text{RNN}}(A,S)$. The RNN discriminator is illustrated in Figure~\ref{fig:rnn_disc}.

The input videos are encoded through two pathways: 1) a pose pathway that encodes the input frames into an N frames by k channels representations about poses $z$ and 2) a motion dynamics pathway that encodes differences of adjacent frames through a 2-layer LSTM-RNN into an (N-1) frames by k channels representation $h$ about motion dynamics. Each frame of $z$ and $h$ are validated using a frame validation module and a diff validation module respectively. The validation scores are averaged over the frames and the final score is a weighted combination of frame score and diff score. The equations of the RNN discriminator is defined in Equation~\ref{eq:RNNDiscriminator}.

\begin{equation}
\footnotesize
\begin{array}{rll}
h_{\text{txt}} &=& \text{LanguageModel}(S) \\
z_i & = & \text{FrameEncoder} (a_i) \triangleq \text{Linear} (a_i),\quad\forall i=1,\dots N \\
d_i & = & \text{DiffEncoder} (a_{i+1},a_i) \triangleq  \text{Linear} (a_{i+1}-a_i),\quad\forall i=1,\dots N-1 \\
h_i, \text{state}_i &=& \text{LSTM}_{\text{diff}} ( d_i,\text{state}_{i-1} ),\quad\forall i=1,\dots N-1\\
z&=&\{z_1,z_2,\dots,z_N\} \\
h&=&\{h_1,h_2,\dots,h_{N-1}\} \\
s_A&=&\text{FrameValidate}(z,h_{\text{txt}}) \\
&\triangleq&\text{Mean} \big( \text{Conv1d} \big\{ \text{ReLU} \lbrack \text{Conv1d} (z)+\text{Conv1d} (h_{\text{txt}}) \rbrack \big\} \big) \\
s_D&=&\text{DiffValidate}(d,h_{\text{txt}}) \\
&\triangleq&\text{Mean} \big( \text{Conv1d} \big\{ \text{ReLU} \lbrack \text{Conv1d} (d)+\text{Conv1d} (h_{\text{txt}}) \rbrack \big\} \big) \\
y&=&w_A s_A + w_D s_D
\end{array}
\label{eq:RNNDiscriminator}
\end{equation}

For experiments, we set the hidden size of all modules to be $k=256$. We pair the RNN discriminator with either the CNN and RNN generators introduced in Section 4.1 and use the same learning protocol as introduced in Section 5.2. Table~\ref{table:rnn_disc} shows the results of the RNN discriminator on the H3.6M dataset, compared with using the CNN discriminator. 
\begin{table*}[t]
\centering
\begin{tabular}{ l c |c c c | c |c c c c }
\toprule
\multicolumn{10}{c}{H3.6M Ablation Study} \\
\midrule
Generator	&Discriminator	& Validation	& Data	& Final & Inception & \multicolumn{4}{c}{Action Retrieval} \\
			&				& 				& Aug. 	& Cut 	& Score & R@1 	& R@3 	& R@5 	& R@10 	\\
\midrule
CNN			& RNN			& 	dense		&	 	&		& 1.87	& 64.7	& 87.4	& 93.4	& 99.3	\\
RNN			& RNN			& 	dense		&	 	&	 	& 1.85	& 70.8	& 86.7	& \textbf{100.0}	& \textbf{100.0}	\\
\midrule
CNN			& CNN			& 	dense		&	 	&		& \textbf{2.53}	& \textbf{97.5}	& \textbf{99.8}	& \textbf{100.0}	& \textbf{100.0} \\
RNN			& CNN			& 	dense		&	 	&	 	& 2.36	& 90.1	& 98.7	& 99.7	& 99.9	\\
\midrule
\multicolumn{4}{c}{Real video samples}				&	 	& 1.48	& 47.2	& 73.9	& 85.2	& 97.0 \\
\bottomrule
\end{tabular}
\caption{Our RNN discriminator results. Comparing the RNN discriminator with our CNN discriminator, the CNN discriminator performs better.}
\label{table:rnn_disc}
\vspace{-28pt}
\end{table*}

Overall, using the CNN discriminator outperforms using the RNN discriminator, for both the CNN and RNN generators. Note that these results are still preliminary. Exploring the space of model architectures for DVGANs is future work.

\section*{Symbol Table}
\begin{tabular}{p{4cm} p{8cm} }
\toprule
Symbol & Description \\
\midrule
Linear\quad & PyTorch torch.nn.Linear, implements $\text{Linear}(x) \triangleq Wx+b$. \\
Conv1d\quad & PyTorch torch.nn.Conv1d, implements the 1-D convolution operation. \\
DnSample\quad & PyTorch torch.nn.AdaptiveAvgPool1d with scale factor 0.5, implements the downsampling by 2 operation. \\
UpSample\quad & PyTorch torch.nn.AdaptiveAvgPool1d with scale factor 2, implements the upsampling by 2 operation. \\
Mean\quad & PyTorch torch.Tensor.mean(dim=0), average the tensor over the time dimension of the animation (dimension 0). \\
Concat(x,y)\quad & Concatenate tensor x and y in the feature dimension (dimension 1). PyTorch torch.Tensor.cat((x,y),dim=1). \\
$\text{out},s_{T+1}=\text{LSTM}(\text{in},s_T)$\quad & An LSTM module that takes input in and current state $s_{T}$, computes output out and next state $s_{T+1}$. Pytorch torch.nn.LSTM. \\
LanguageModel(S) & A language model that embeds words in word embeddings, and then use the word embeddings as inputs to an LSTM. The hidden and cell states after reading in the last word are concatenated as the representation of the sentence S. The word embeddings are learned from scratch in our work. \\ 
\bottomrule
\end{tabular}

\begin{figure*}
\centering
\includegraphics[width=\textwidth]{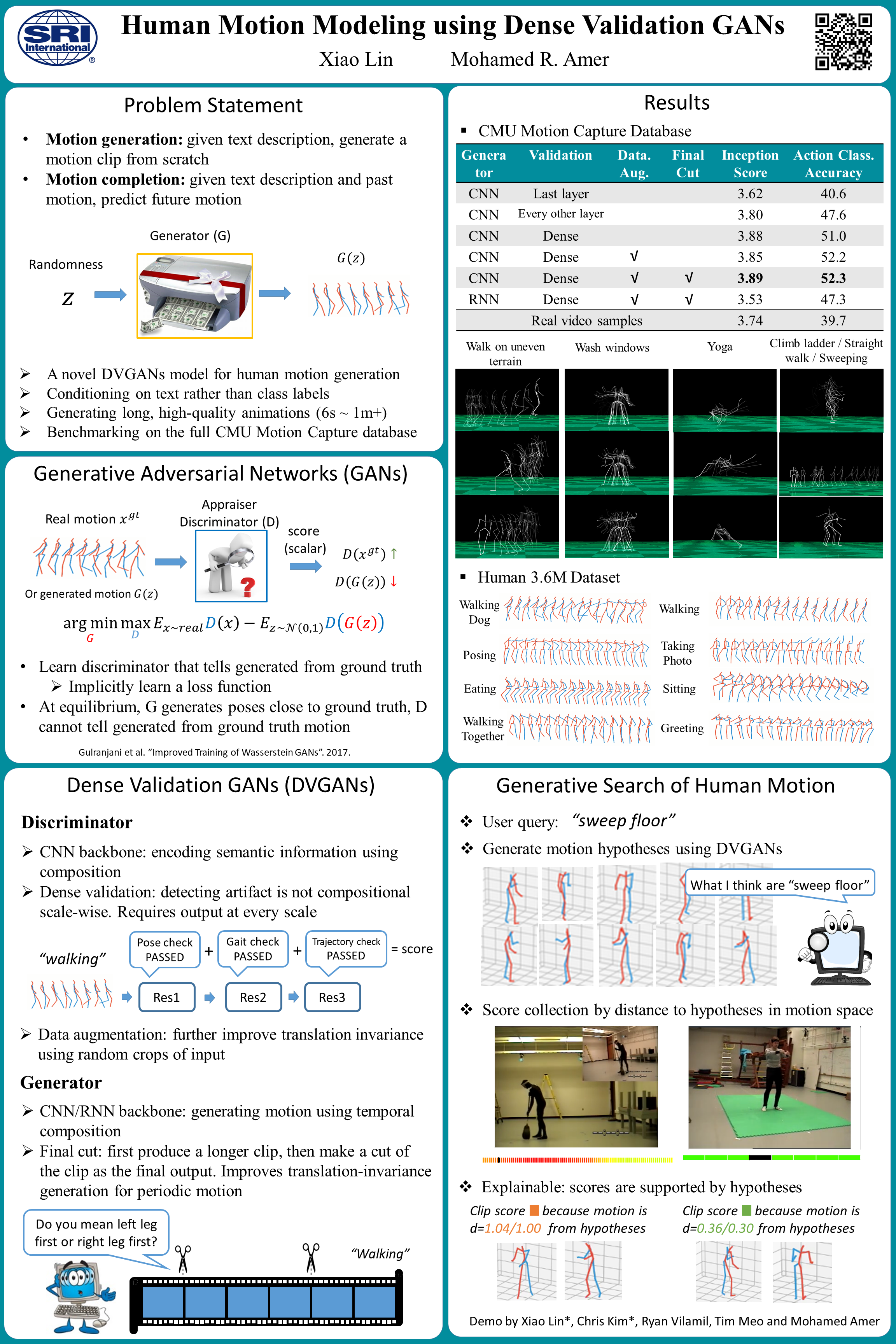}
\label{fig:poster}
\end{figure*}

\end{document}